\pdfoutput=1
\documentclass[sigconf]{acmart}
\usepackage{multirow}
\AtBeginDocument{%
  }

\copyrightyear{2026}
\acmYear{2026}
\setcopyright{cc}
\setcctype{by}
\acmConference[MM '26]{Proceedings of the 34th ACM International Conference on Multimedia}{November 10--14, 2026}{Rio de Janeiro, Brazil}
\acmBooktitle{Proceedings of the 34th ACM International Conference on Multimedia (MM '26), November 10--14, 2026, Rio de Janeiro, Brazil}
\acmDOI{10.1145/3767308.3837699}
\acmISBN{979-8-4007-2213-4/2026/11}

\begin{document}

\title{Evidence-Guided Detection, Localization and Explanation for Text-Centric Image Forensics}

\author{Peifeng Liu}
\affiliation{%
  \institution{Guangdong Provincial Key Laboratory of Intelligent Information Processing, Shenzhen Key Laboratory of Media Security}
  \institution{Shenzhen University}
  \city{Shenzhen}
  \country{China}
}
\email{2410043004@mails.szu.edu.cn}

\author{Bin Li}
\authornote{Corresponding author.}
\affiliation{%
  \institution{Guangdong Provincial Key Laboratory of Intelligent Information Processing, Shenzhen Key Laboratory of Media Security, SZU-AFS Joint Innovation Center for AI Technology}
  \institution{Shenzhen University}
  \city{Shenzhen}
  \country{China}
}
\email{libin@szu.edu.cn}

\author{Qingsong Zhang}
\affiliation{%
  \institution{Guangdong Provincial Key Laboratory of Intelligent Information Processing, Shenzhen Key Laboratory of Media Security}
  \institution{Shenzhen University}
  \city{Shenzhen}
  \country{China}
}
\email{2230707827@qq.com}

\author{Yangxin Yu}
\affiliation{%
  \institution{Guangdong Provincial Key Laboratory of Intelligent Information Processing, Shenzhen Key Laboratory of Media Security}
  \institution{Shenzhen University}
  \city{Shenzhen}
  \country{China}
}
\email{2453043007@mails.szu.edu.cn}

\author{Leqing Chen}
\affiliation{%
  \institution{Guangdong Provincial Key Laboratory of Intelligent Information Processing, Shenzhen Key Laboratory of Media Security}
  \institution{Shenzhen University}
  \city{Shenzhen}
  \country{China}
}
\email{2400042018@mails.szu.edu.cn}

\author{Xiaoye Qiu}
\affiliation{%
  \institution{Guangdong Provincial Key Laboratory of Intelligent Information Processing, Shenzhen Key Laboratory of Media Security}
  \institution{Shenzhen University}
  \city{Shenzhen}
  \country{China}
}
\email{2553043008@mails.szu.edu.cn}

\renewcommand{\shortauthors}{Peifeng Liu, Bin Li, Qingsong Zhang, Yangxin Yu, Leqing Chen, and Xiaoye Qiu}

\begin{abstract}
  The rapid progress of AIGC has made text-centric image manipulation increasingly accessible, creating new forensic challenges that require not only authenticity detection but also spatial grounding and evidence-based explanation. This paper presents our solution to the GenText-Forensics Challenge at ACM Multimedia 2026. We propose an evidence-guided detector-localizer-reasoner system, where an image-level detector provides a global authenticity prior, a dedicated localizer extracts tampered regions as spatial grounding evidence, and an MLLM-based reasoner generates structured forensic reports grounded in this expert forensic evidence. These modules are connected through a cascaded evidence flow: the detector gates the subsequent localization and prompting process, the localizer converts tamper responses into grounding boxes, and the reasoner is trained to synthesize the detector decision and localized evidence into the final report. As a key part of our method, we introduce iterative difficulty-aware mining to improve localization quality and apply report-mask consistency post-processing to align report grounding with predicted masks. On the official hidden test set, our system achieves a final score of 0.638 and ranks second in the challenge, validating the effectiveness of the proposed evidence-guided system. The code is available at \href{https://github.com/peifengLiu42/ACMMM26-evidence-guided-detector-localizer-reasoner-system}{this https URL}.
\end{abstract}



\maketitle

\section{Introduction}
Text-centric images like documents and receipts serve as key evidence in daily work. Advanced editing tools \cite{gpt4o,gemini,bagel,seededit,flux2024,qwenimage,step1x} enable stealthy text tampering that retains original layouts but modifies critical information such as amounts and timestamps. Tiny altered areas can completely distort evidentiary validity. Thus, text image forensics requires not only authenticity classification but also tampering localization and evidence-based analysis reports.

The GenText-Forensics Challenge at ACM Multimedia 2026 formulates text-centric image forensics as a unified forensic report generation task based on the RealText-V2 dataset. Given a text-rich image, participants are required to produce a structured Markdown-style report that integrates authenticity judgment, spatial grounding, and evidence-based reasoning. The evaluation jointly considers detection, localization, explanation, and report quality.

Unlike natural image manipulation~\cite{casia,columbia,coverage,dso,imd2020} or conventional low-resolution document tampering detection~\cite{DTD}, the RealText-V2 setting focuses on high-resolution text-centric images with dense textual content, complex layouts, and strong semantic dependencies. In this setting, forgery evidence is inherently heterogeneous. Image-level cues are encoded in global feature representations that capture long-range correlations and document-wide statistical deviations, providing a holistic authenticity prior. Region-level traces arise from local edits around modified characters or pasted text, such as edge blending defects, background discontinuity, and compression mismatch, requiring precise spatial localization. Semantic-level anomalies, such as altered words, numbers, or fields, may further introduce logical contradictions or misleading meanings that require contextual interpretation.

These heterogeneous evidence types differ in granularity and feature space, making it difficult for a single model to handle all forensic requirements. Dedicated forensic models~\cite{CAT-Net,Trufor,sparsevit,DTD,ffdn,ascformer} are effective at capturing low-level tampering traces and producing detection scores, heatmaps, or masks, but they generally lack semantic reasoning and structured report-generation capability. In contrast, direct Multimodal Large Language Model (MLLM)~\cite{Qwen3-vl,llavanext,internvl3,gemini,gpt4o} inference provides strong semantic understanding and language reasoning, but remains limited in fine-grained visual perception for subtle artifact analysis and precise localization.

Guided by these insights, we propose an evidence-guided detector-localizer-reasoner framework that decomposes forensic tasks into specialized modules and aggregates their outputs into a unified report. For image-level judgment, a detector with auxiliary localization supervision learns global authenticity priors while retaining tampering-sensitive spatial cues. For region-level localization, we present Difficulty-Aware Tamper Localization, which iteratively mines hard samples and error-prone regions, then fine-tunes the model via difficulty-weighted sampling and hard-region cropping. This strategy enhances localization performance on small edits and ambiguous boundaries in high-resolution text-centric forgeries.

Powered by the detector's authenticity prediction and suspicious-region bounding boxes, the LLM reasoner generates a structured forensic report via an evidence-conditioned prompt. Unlike direct MLLM inference, which may produce ungrounded claims from semantic priors, our reasoner is strictly constrained by expert module outputs, reducing hallucinations and aligning authenticity judgment, spatial localization, and textual analysis. Our system achieved second place in the final evaluation, validating the effectiveness of our design.

Overall, our system offers three main advantages:
\begin{itemize}
\item \textbf{Complementary Expert-LLM Collaboration.} Specialized visual models handle detection and localization, while the LLM-based reasoner performs semantic integration and report generation.
\item \textbf{Difficulty-Aware Spatial Evidence Enhancement.} The localization strategy mines hard samples and error-centered regions, improving spatial grounding for report generation.
\item \textbf{Evidence-Grounded Report Generation.} Conditioning the reasoner on detector decisions and localized regions improves alignment between generated reports and expert forensic evidence.
\end{itemize}

\section{Related Work}

Image manipulation detection and localization (IMDL) aims to identify manipulated images and localize tampered regions. Existing methods can be broadly divided into expert forensic models and MLLM-centric models. The former focus on low-level manipulation traces, while the latter introduce multimodal semantic priors for interpretable reasoning.

\textbf{Expert Models.}
Expert forensic models detect manipulation by modeling visual artifacts such as boundary inconsistency, noise mismatch, compression traces, and statistical anomalies. For general image manipulation detection, representative methods exploit RGB-DCT artifacts~\cite{CAT-Net}, noise and boundary cues~\cite{MVSS}, noise-sensitive fingerprints~\cite{Trufor}, or non-semantic manipulation features~\cite{sparsevit}. For document image forensics, DTD~\cite{DTD} and FFDN~\cite{ffdn} incorporate DCT, wavelet, and frequency-domain cues to improve robustness against post-processing attacks. TFAM~\cite{dong2024robust} fuses local-global contextual information, while CD-SD~\cite{cdsd} and DITL\textsuperscript{2}~\cite{DITL2} address cross-domain robustness through color-semantic decoupling and frequency-domain adaptation. These methods are effective at producing detection scores, heatmaps, or masks, but generally lack semantic reasoning and structured report-generation capability.

\textbf{MLLM-Centric Models.}
Recent studies have introduced MLLMs into forgery analysis to enhance semantic understanding and interpretability. Methods such as FakeShield~\cite{fakeshield}, ForgeryGPT~\cite{forgerygpt}, SIDA~\cite{sida}, and ForgerySleuth~\cite{forgerysleuth} combine MLLMs with domain-aware modules, mask-aware extractors, detection/segmentation tokens, or trace encoders to connect low-level forgery cues with high-level reasoning. For text- and document-centric forensics, TVSIP~\cite{tvsip} integrates expert-extracted visual artifacts with MLLM-identified semantic conflicts, while VeriChain~\cite{verichain} explores reasoning-chain-based verification. These works show the potential of MLLMs for interpretable forensics. Following this direction, our method further treats detector outputs and localized tamper regions as explicit forensic evidence to condition MLLM reasoning, enabling the final report to be grounded in both global authenticity judgment and spatial localization evidence.

\section{Proposed Solution}

\begin{figure*}[t]  
  \centering
  \includegraphics[width=\linewidth]{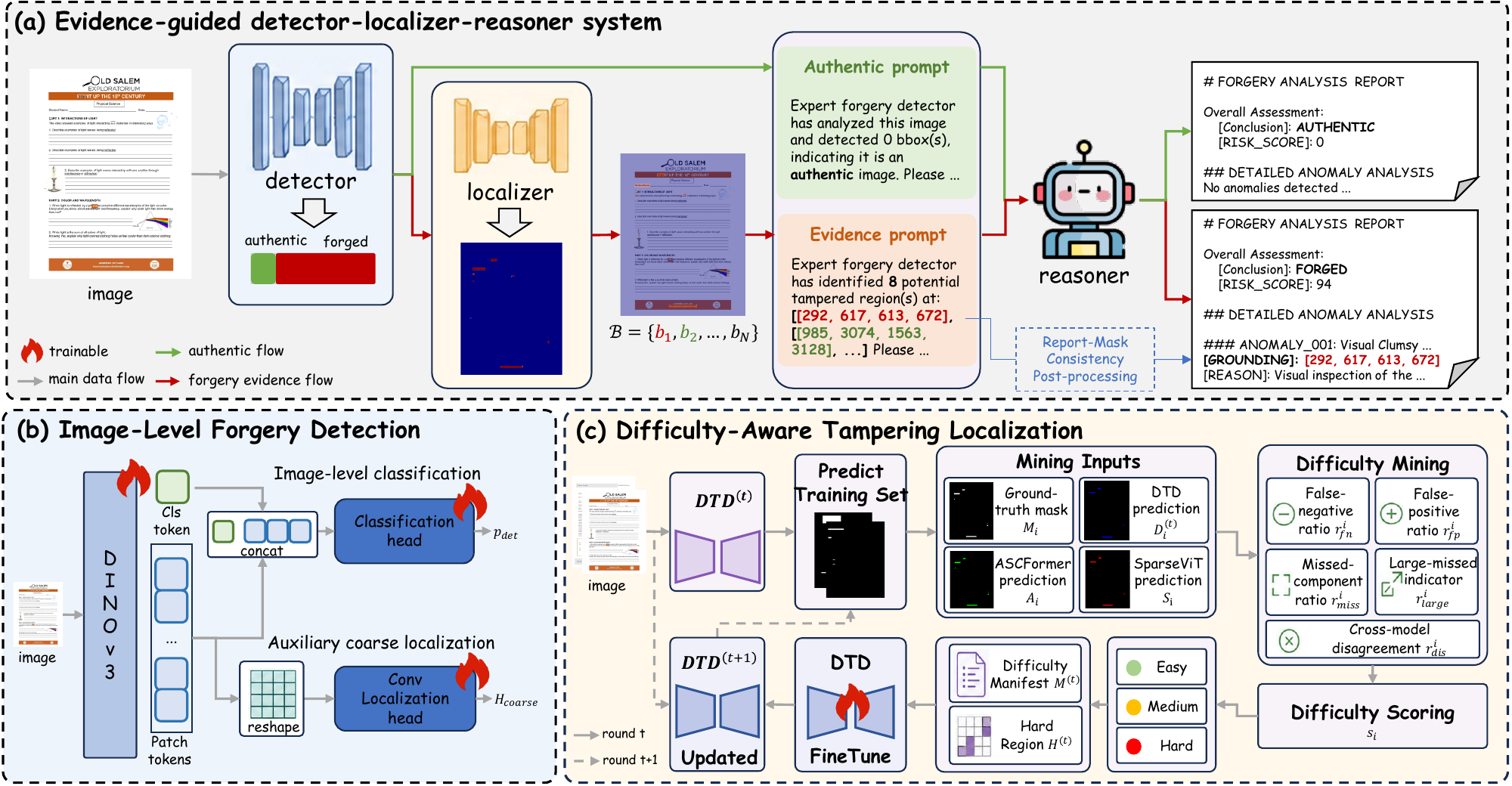}
  \caption{Overview of the proposed solution. (a) The evidence-guided detector-localizer-reasoner system for forensic report generation. (b) The image-level forgery detector, where DINOv3 tokens support both authenticity classification and auxiliary coarse localization. (c) The iterative difficulty-aware localization process, which mines hard samples and error-centered crops to improve tamper grounding.}
  \Description{Pipeline overview of the evidence-guided detector-localizer-reasoner system, including image-level detection, localization, and iterative difficulty-aware mining.}
  \label{fig:method}
\end{figure*}

\subsection{Overview}

As illustrated in Fig.~\ref{fig:method} (a), we propose an evidence-guided detector-localizer-reasoner system for text-centric image forensic report generation. Given an input image $I$, the detector first predicts an image-level forgery probability:
\begin{equation}
p_{\mathrm{det}}=\mathcal{D}(I),
\end{equation}
where $\mathcal{D}(\cdot)$ denotes the detector. The predicted label $\hat{y}$ is obtained by thresholding $p_{\mathrm{det}}$ with $\tau_{\mathrm{det}}$, where $\hat{y}=\mathrm{forged}$ if $p_{\mathrm{det}}\geq\tau_{\mathrm{det}}$, and $\hat{y}=\mathrm{authentic}$ otherwise. This detector output serves as a global expert prior and controls whether spatial evidence should be further extracted.

When the image is predicted as forged, the localizer is activated to produce a tamper probability heatmap $H=\mathcal{F}(I)$. The heatmap is then binarized, filtered by connected components, and converted into bounding boxes:
\begin{equation}
\mathcal{B}=
\left\{
\mathrm{BBox}(c)
\mid
c\in \mathrm{CC}\bigl(\mathbb{I}(H\geq \tau_{\mathrm{loc}})\bigr),
\ \mathrm{Area}(c)\geq \alpha
\right\},
\end{equation}
where $\tau_{\mathrm{loc}}$ is the localization threshold and $\alpha$ is the minimum area threshold. The resulting bounding boxes $\mathcal{B}$ serve as explicit spatial expert evidence for the reasoning stage.

The prompt for the MLLM-based reasoner is constructed in a detector-gated manner:
\begin{equation}
\mathcal{P} =
\begin{cases}
\mathcal{P}_{\mathrm{evi}}(I,\hat{y},\mathcal{B}),
& \hat{y}=\mathrm{forged},\ |\mathcal{B}|>0, \\[1mm]
\mathcal{P}_{\mathrm{auth}}(I,\hat{y}),
& \mathrm{otherwise}.
\end{cases}
\end{equation}
Here, $\mathcal{P}_{\mathrm{evi}}$ denotes an evidence-conditioned prompt containing localized suspicious regions, while $\mathcal{P}_{\mathrm{auth}}$ denotes an authentic-oriented prompt without forced grounding regions. Thus, forged-oriented reasoning is triggered only when both the detector prediction and localized expert evidence support it.

The initial report is generated by the reasoner:
\begin{equation}
R_{0}=\mathcal{R}(\mathcal{P}),
\end{equation}
where $\mathcal{R}(\cdot)$ denotes the MLLM-based reasoner. To further enforce consistency between report grounding and localization masks, we apply report-mask consistency post-processing:
\begin{equation}
R=\Psi_{\mathrm{rmc}}(R_{0},\mathcal{B}),
\end{equation}
where $\Psi_{\mathrm{rmc}}(\cdot)$ refines the \texttt{[GROUNDING]} fields in the generated report using localization-derived boxes and supplements missing localized regions when necessary. 

In this way, the final report is constrained by expert-module evidence from both detection and localization, improving the consistency among authenticity judgment, spatial grounding, and evidence-based explanation.

\subsection{Auxiliary Localization-Supervised Image-Level Forgery Detection}

As illustrated in Fig.~\ref{fig:method} (b), the detector provides a global authenticity prior for the whole pipeline. We adopt DINOv3~\cite{dinov3} as the visual backbone because its transformer representations can capture long-range visual dependencies in high-resolution text-centric images, while its patch tokens also preserve spatial information useful for tampering-aware supervision. Rather than using the CLS token alone for classification, we concatenate it with the mean-pooled patch tokens to combine the global summary and distributed local responses:
\begin{equation}
\mathbf{z}_{\mathrm{img}}
=
\left[
\mathbf{z}_{\mathrm{cls}};
\frac{1}{N}\sum_{i=1}^{N}\mathbf{z}_{i}
\right],
\end{equation}
where $\mathbf{z}_{\mathrm{cls}}$ is the CLS token and $\{\mathbf{z}_{i}\}_{i=1}^{N}$ are the patch tokens. The forged probability is then predicted by a lightweight classification head:
\begin{equation}
p_{\mathrm{det}}=\sigma\left(f_{\mathrm{cls}}(\mathbf{z}_{\mathrm{img}})\right).
\end{equation}

A classification-only detector may learn image-level separability without preserving spatially discriminative tampering cues. Therefore, we attach an auxiliary convolutional localization head to the patch tokens. The patch tokens are reshaped into a two-dimensional feature map and fed into a lightweight convolutional decoder to predict a coarse tampering heatmap:
\begin{equation}
H_{\mathrm{coarse}}
=
f_{\mathrm{conv}}\left(\mathrm{Reshape}\left(\{\mathbf{z}_{i}\}_{i=1}^{N}\right)\right).
\end{equation}
The detector is trained with a joint objective:
\begin{multline}
\mathcal{L}_{\mathrm{det}}
= \lambda_{\mathrm{1}}\mathcal{L}_{\mathrm{BCE}}(p_{\mathrm{det}}, y)
+ \lambda_{\mathrm{2}}\mathcal{L}_{\mathrm{CE}}(H_{\mathrm{coarse}}, M) \\
+ \lambda_{\mathrm{3}}\mathcal{L}_{\mathrm{Dice}}(H_{\mathrm{coarse}}, M),
\end{multline}
where $y$ is the image-level label and $M$ is the ground-truth tampering mask. The auxiliary localization head is used to regularize the detector with pixel-level supervision, encouraging the shared backbone to retain tampering-sensitive spatial cues.

\begin{table*}[!t]
  \centering
\caption{Comparison with state-of-the-art methods on the internal validation split. We report detection, localization, explanation, and the reproducible official weighted score. ``-'' denotes unavailable metrics. SIDA$^*$ and TVSIP$^*$ denote SIDA~\cite{sida} and TVSIP~\cite{tvsip} implemented with Qwen3-VL-4B~\cite{Qwen3-vl} as the MLLM. Bold indicates the best performance.}

  \label{tab:main_results}
  \begin{tabular}{c *{9}{c}}
    \toprule
    \multirow{2}{*}{\textbf{Method}}
    & \multicolumn{2}{c}{\textbf{Detection}}
    & \multicolumn{4}{c}{\textbf{Localization}}
    & \multicolumn{2}{c}{\textbf{Explanation}}
    & \multirow{2}{*}{\textbf{Score}} \\
    \cmidrule(lr){2-3} \cmidrule(lr){4-7} \cmidrule(lr){8-9}
    & Bal-ACC & Wtd-F1
    & P & R & F1 & IoU
    & CSS & BS
    & \\
    \midrule
    Trufor    & 94.60 & 94.67 & 78.26 & 63.95 & 66.80 & 54.33 & - & - & - \\
    SparseViT & 93.67 & 93.64 & 63.30 & 38.92 & 44.93 & 33.16 & - & - & - \\
    DINOv3-Conv & 99.57 & \textbf{99.56} & 21.08 & \textbf{90.26} & 30.81 & 20.41 & - & - & - \\
    DINOv3-Mask2Former & 99.18 & 99.11 & 42.79 & 88.53 & 54.77 & 40.58 & - & - & - \\
    Swin-Mask2Former & 87.97 & 86.72 & 21.39 & 83.65 & 31.96 & 20.51 & - & - & - \\
    DTD         & 95.08 & 95.53 & 83.05 & 70.34 & 72.49 & 61.58 & - & - & - \\
    ASCFormer   & 91.15 & 91.84 & 81.51 & 60.96 & 65.16 & 53.11 & - & - & - \\
    \midrule
    InternVL3.5-4B & 96.20 & 96.29 & 23.65 & 23.17 & 21.68 & 14.97 & 90.86 & 83.96 & 47.47 \\
    Qwen3VL-2B     & 95.40 & 95.48 & 52.94 & 24.27 & 29.37 & 19.14 & 96.48 & 84.68 & 49.00 \\
    Qwen3VL-4B     & 93.92 & 93.79 & 41.38 & 38.80 & 37.28 & 26.35 & 96.40 & 84.26 & 51.32 \\
    Qwen3VL-8B     & 98.15 & 98.15 & 64.22 & 32.04 & 38.74 & 26.75 & \textbf{96.61} & \textbf{87.49} & 53.27 \\
    \midrule
    SIDA$^*$        &   96.95    &   97.04    &   60.63    &  35.63     &   41.73    &  30.53     &   89.21    &   83.58    &  53.86     \\
    TVSIP$^*$       & 91.85 & 92.58 & 54.53 & 68.13 & 56.65 & 43.21 & 96.20 & 83.77 & 57.62 \\
    \midrule
    Ours            &   \textbf{99.60}    &   \textbf{99.56}    &   \textbf{83.71}    &   79.61    &   \textbf{79.03}    &   \textbf{69.93}    &   91.01    &   80.15    &   \textbf{69.86}    \\
    \bottomrule
  \end{tabular}
\end{table*}

\subsection{Difficulty-Aware Tampering Localization}
The localization branch is activated for images classified as forged by the detector, aiming to identify tampered regions for subsequent evidence-conditioned reasoning. However, the difficulty of localization varies significantly across high-resolution text-centric images. After initial training, many easy samples can already be localized reliably and provide limited additional supervision if repeatedly sampled. In contrast, medium and hard samples, such as small text edits, ambiguous boundaries, confusing backgrounds, and regions with strong model disagreement, are less frequent but contain more informative error signals. Uniform sampling therefore allocates many updates to already learned patterns while insufficiently revisiting the localizer's failure cases. This motivates an iterative difficulty-aware mining strategy that converts the current model's localization errors into targeted fine-tuning samples.

As shown in Fig.~\ref{fig:method} (c), we adopt DTD~\cite{DTD} as the primary localizer, since it is designed for document tampering localization and captures both visual artifacts and frequency-domain inconsistencies. Given an image $I$, DTD predicts a tampering probability map:
\begin{equation}
H=\mathcal{F}_{\mathrm{dtd}}(I),
\end{equation}
and is optimized with pixel-level mask supervision:
\begin{align}
\mathcal{L}_{\mathrm{loc}}
&=
\lambda_{4}\mathcal{L}_{\mathrm{CE}}(H,M)
+
\lambda_{5}\mathcal{L}_{\mathrm{Dice}}(H,M),
\end{align}
where $M$ denotes the ground-truth tampering mask. We first train a baseline DTD localizer on the RealText-V2 training set. This model serves as the initial state $\mathcal{F}_{\mathrm{dtd}}^{(0)}$ for difficulty mining, so that hard samples are defined according to actual localization failures rather than hand-crafted assumptions.

At mining round $t$, let $I_i$ denote the $i$-th full image in the training set and $\mathcal{F}_{\mathrm{dtd}}^{(t)}$ denote the current DTD localizer. We run $\mathcal{F}_{\mathrm{dtd}}^{(t)}$ on $I_i$ and obtain its binary prediction $D_i^{(t)}$. To capture complementary error cues, we also use predictions from ASCFormer~\cite{ascformer} and SparseViT~\cite{sparsevit}, denoted as $A_i$ and $S_i$, respectively. Thus, the mining input for each image is:
\begin{equation}
\{M_i, D_i^{(t)}, A_i, S_i\},
\end{equation}
The mining is conducted at the full-image level, while the extracted error regions are later used to guide crop selection during fine-tuning.

Taking $M_i$ as reference, we compute the DTD false-negative and false-positive regions:
\begin{equation}
FN_i^{(t)} = M_i \setminus D_i^{(t)},
\qquad
FP_i^{(t)} = D_i^{(t)} \setminus M_i .
\end{equation}
We then summarize each image by a difficulty score that combines pixel-level errors, component-level missed regions, large-region failures, and cross-model disagreement:
\begin{align}
s_i^{(t)}
&=
0.35 r_{\mathrm{fn}}^i
+
0.20 r_{\mathrm{miss}}^i
+
0.15 r_{\mathrm{large}}^i \notag\\
&\quad+
0.15 \min(r_{\mathrm{fp}}^i,1)
+
0.15 r_{\mathrm{dis}}(D_i^{(t)},A_i,S_i) ,
\end{align}
where $r_{\mathrm{fn}}^i$ and $r_{\mathrm{fp}}^i$ measure false-negative and false-positive ratios, $r_{\mathrm{miss}}^i$ measures missed connected components, $r_{\mathrm{large}}^i$ indicates whether a large tampered component is missed, and $r_{\mathrm{dis}}(D_i^{(t)},A_i,S_i)=1-\mathrm{IoU}(D_i^{(t)}, A_i\cup S_i)$ measures disagreement between DTD and the auxiliary localizers.

Based on $s_i^{(t)}$ and the localization statistics, each image is assigned to an easy, medium, or hard group using empirically defined rule-based criteria. These criteria are designed according to the main failure modes observed in the baseline localizer: severe missed tampered regions, excessive false positives, low localization quality, and large connected components being missed.

\begin{equation}
g_i =
\begin{cases}
\mathrm{hard},
& (s_i^{(t)} \geq 0.60) \vee (F_i < 0.40) \vee (r_{\mathrm{fn}}^i > 0.45) \\
&\quad \vee (r_{\mathrm{fp}}^i > 0.80) \vee (r_{\mathrm{miss}}^i > 0.50) \vee (r_{\mathrm{large}}^i = 1),\\[1mm]
\mathrm{easy},
& (s_i^{(t)} < 0.25) \wedge (F_i \geq 0.75) \wedge (R_i \geq 0.80) \\
&\quad \wedge (r_{\mathrm{fn}}^i \leq 0.15) \wedge (r_{\mathrm{fp}}^i \leq 0.20) \wedge (r_{\mathrm{large}}^i = 0),\\[1mm]
\mathrm{medium},
& \mathrm{otherwise}.
\end{cases}
\end{equation}
Here, $F_i$ and $R_i$ denote the DTD F1-score and recall on image $I_i$, respectively. This rule treats samples with severe localization failures as hard, regards only consistently well-localized samples as easy, and assigns the remaining ambiguous cases to the medium group.

The mining process produces an image-level difficulty manifest $\mathcal{M}^{(t)}=\{(I_i,s_i^{(t)},g_i)\}$ and a region-level hard set $\mathcal{H}^{(t)}=\{(I_i,b_i^j,q_i^j)\}$, where $b_i^j$ is the bounding box of an error component and $q_i^j$ denotes its error type, such as false negative, false positive, auxiliary-recovered false negative, or all-model-missed region. During fine-tuning, full images are sampled according to fixed group probabilities:
\begin{equation}
\pi_{\mathrm{easy}}=0.05,\qquad
\pi_{\mathrm{medium}}=0.35,\qquad
\pi_{\mathrm{hard}}=0.60.
\end{equation}
After a full image is selected, a $512\times512$ crop is extracted; for hard samples, the crop is centered around a mined hard region with probability $0.75$, with random jitter applied to avoid repeatedly using the same crop.
In each mining-and-fine-tuning stage, we fine-tune the localizer for 30 epochs using $\mathcal{M}^{(t)}$ and $\mathcal{H}^{(t)}$, and select the best checkpoint for the next mining stage. The updated localizer is used to re-predict the training set and refresh the hard samples:
\begin{equation}
\mathcal{F}_{\mathrm{dtd}}^{(t+1)}
=
\mathrm{FineTune}
\left(
\mathcal{F}_{\mathrm{dtd}}^{(t)};
\mathcal{M}^{(t)},\mathcal{H}^{(t)}
\right).
\end{equation}
This closed-loop procedure allows the localizer to progressively shift its learning focus from the initial failure cases to the remaining hard regions that are still poorly localized after each model update.

\subsection{SFT for the Evidence-Conditioned Reasoner}
To make the reasoner follow the evidence provided by expert modules, we construct QA-style SFT data from RealText-V2 image-mask-report triplets. For each image, we convert the tampering mask into normalized bounding boxes in the $[0,999]$ range. Authentic samples are assigned zero boxes, while forged samples include the extracted suspicious boxes in the question. The question $Q_i$ therefore contains the input image and an evidence-aware instruction, asking the reasoner to generate a forensic report based on the provided detector/localizer evidence. The answer $A_i$ is the challenge-provided reference forensic report, with grounding coordinates normalized to the same format. The resulting SFT dataset is defined as:
\begin{equation}
\mathcal{D}_{\mathrm{sft}}=\{(Q_i,A_i)\}_{i=1}^{N}.
\end{equation}
We fine-tune the reasoner with the standard autoregressive objective:
\begin{equation}
\mathcal{L}_{\mathrm{sft}}
=
-\sum_{i,t}\log p_{\theta}(a_{i,t}\mid Q_i,a_{i,<t}),
\end{equation}
where $a_{i,t}$ denotes the $t$-th token of the target report.

\subsection{Report-Mask Consistency Post-processing}

To improve spatial consistency, we refine the generated report after inference using the final localization mask. Specifically, we parse the report \texttt{[GROUNDING]} boxes and replace each one with the most overlapping box extracted from the localization mask.

If the localization mask contains additional boxes not covered by the report, we append them as supplementary anomaly entries. These entries only include localization-derived grounding coordinates, with the reason field marked as an expert-module supplement. This preserves the original analysis while ensuring better consistency between the report and the predicted mask.

\section{Experiments}

\subsection{Experimental Setup}

\textbf{Dataset.}
We conducted experiments on the official RealText-V2 dataset from the GenText-Forensics Challenge. Since the official test annotations were withheld, we constructed an internal validation split by randomly sampling 10\% of real images and 10\% of forged images from the training set, with a fixed random seed of 42 for reproducibility; the remaining data was used for training. This split was used for component-level evaluation before submitting the complete system to the official evaluation server.

\textbf{State-of-the-Art Methods.}
To comprehensively evaluate our framework, we compared it with representative methods from three categories. For expert forensic models, we included general IMDL methods, i.e., TruFor \cite{Trufor} and SparseViT \cite{sparsevit}, as well as segmentation-based and document-oriented baselines, including DINOv3-Mask2Former~\cite{dinov3,mask2former}, Swin-Mask2Former~\cite{swin,mask2former}, DTD \cite{DTD}, and ASCFormer \cite{ascformer}. For MLLM-centric forensic models, we included SIDA \cite{sida} and TVSIP~\cite{tvsip}. We further evaluated general-purpose MLLMs, including Qwen3-VL \cite{Qwen3-vl} and InternVL3.5 \cite{internvl3}.

\textbf{Evaluation Metrics.}
We assess three dimensions: detection, localization and explanation. Detection metrics include balanced accuracy (Bal-Acc) and weighted F1 (Wtd-F1); localization uses pixel-wise precision, recall, F1 and IoU; explanation is measured via BERTScore (BS)~\cite{bertscore} and cosine semantic similarity (CSS). We unify metric calculation for fair baseline comparison. For expert models, detection results use the classification head output with threshold 0.5. Models lacking a classification branch predict authenticity from the maximum value of segmentation masks (threshold 0.5). For MLLM-centric models, we parse authenticity labels and coordinates from generated text, then convert coordinates to binary masks for localization evaluation. The official subjective LLM-based report score is unstable in internal tests. We thus calculate a reproducible objective partial score using Detection F1, Localization IoU and BERTScore, weighted by official coefficients 0.3, 0.4 and 0.15.

\begin{table}[t]
  \centering
\caption{Ablation study of key components. We evaluate evidence guidance, auxiliary localization loss, and report-mask consistency post-processing.}

  \label{tab:ablation_evidence}
  \resizebox{\linewidth}{!}{
  \begin{tabular}{c cc cc cc}
    \toprule
    \multirow{2}{*}{\textbf{Ablation}} 
    & \multicolumn{2}{c}{\textbf{Detection}} 
    & \multicolumn{2}{c}{\textbf{Localization}} 
    & \multicolumn{2}{c}{\textbf{Explanation}} \\
    \cmidrule(lr){2-3} \cmidrule(lr){4-5} \cmidrule(lr){6-7}
    & \textbf{Bal-ACC} & \textbf{Wtd-F1} 
    & \textbf{F1} & \textbf{IoU} 
    & \textbf{CSS} & \textbf{BS} \\
    \midrule
    Baseline              & 93.92 & 93.79 & 37.28 & 26.35 & 96.40 & 84.26 \\
    \textit{w/o} Detector & 95.70  & 95.81  &   79.35    &   70.21    &   90.89    &  80.17     \\
    \textit{w/o} Localizer& 99.57  &  99.56  &  44.65  &  32.33 & 91.05  &  81.61 \\
    \textit{w/o} Aux. Loc. Loss &   97.27    &   97.04    &   76.30    &  67.68  &  91.00  &  80.08  \\
    \textit{w/o} RMC pp.  & 99.40 & 99.33 & 62.98 & 49.67 & 91.01 & 80.20 \\
    Full Model            & 99.60 & 99.56 & 79.03 & 69.93 & 91.01 & 80.15 \\
    \bottomrule
  \end{tabular}
  }
\end{table}

\begin{table}[t]
\centering
\caption{Effect of difficulty-aware tampering localization.}
\label{tab:abl_difficulty_mining}
\resizebox{\linewidth}{!}{
    \begin{tabular}{c ccccc}
    \toprule
    \textbf{Ablation} & \textbf{P} & \textbf{R} & \textbf{F1} & \textbf{IoU} & \textbf{$\Delta$IoU} \\
    \midrule
    DTD Baseline                & 78.32 & 72.38 & 71.67 & 60.40 & -- \\
    Difficulty Mining (Round 1) & 82.40 & 77.98 & 77.32 & 67.38 & +6.98 \\
    Difficulty Mining (Round 2) & 83.71 & 79.61 & 79.03 & 69.93 & +9.53 \\
    Difficulty Mining (Round 3) & 81.17 & 76.60 & 75.85 & 65.54 & +5.14 \\
    \bottomrule
    \end{tabular}
}
\end{table}

\textbf{Implementation Details.}
All experiments were conducted on four NVIDIA L40 GPUs. For fair comparison, all trainable baselines and our framework were trained on the same training split and evaluated on the same internal validation split. For the detector, we fine-tuned DINOv3 ViT-B/16~\cite{dinov3} at an input resolution of $896 \times 1344$ for 12 epochs with a batch size of 12, using a backbone learning rate of $1\times10^{-5}$ and a head learning rate of $1\times10^{-4}$. For the localizer, we used DTD~\cite{DTD} trained with the proposed Difficulty-Aware Tamper Localization strategy for two mining-and-fine-tuning rounds, i.e., $t=2$. For the reasoner, we fine-tuned Qwen3-VL-4B-Instruct~\cite{Qwen3-vl} via LoRA~\cite{lora} SFT for 20 epochs. We froze the visual encoder, enabled the MM projector, and only applied LoRA adapters to the LLM backbone, with LoRA rank set to 32 and alpha set to 64.

\subsection{Main Results}

Table~\ref{tab:main_results} presents the overall comparison with expert forensic models, general-purpose MLLMs, and MLLM-centric forensic methods. Our method achieves the best overall score, indicating that the proposed evidence-guided system effectively balances image-level detection, pixel-level localization, and explanation quality. The results show that expert models are generally stronger than general-purpose MLLMs in tamper localization, while MLLMs provide competitive semantic explanation ability but lack reliable fine-grained grounding. By combining detector-derived authenticity priors, localization-based spatial evidence, and evidence-conditioned reasoning, our method obtains more balanced performance across all evaluation dimensions. These results validate the effectiveness of using expert forensic evidence to guide MLLM-based report generation for text-centric image forensics.

\subsection{Ablation Studies}

\textbf{Effect of Evidence Guidance.}
Table~\ref{tab:ablation_evidence} shows that explicit expert evidence is crucial for balanced forensic report generation. The baseline, which relies mainly on direct MLLM reasoning, obtains weak localization performance despite reasonable explanation scores. Removing the detector weakens image-level authenticity judgment, while removing the localizer causes a clear drop in spatial grounding, showing that global and regional evidence play complementary roles. The full model achieves the best overall balance by combining detector-derived authenticity priors with localization-based spatial evidence, enabling the reasoner to generate reports grounded in expert forensic evidence.

\textbf{Effect of Auxiliary Localization Loss.}
As shown in Table~\ref{tab:ablation_evidence}, removing the auxiliary localization loss from the detector degrades both detection and localization-related performance. This indicates that image-level classification alone is insufficient for learning tampering-sensitive representations. By introducing coarse mask supervision, the detector is encouraged to preserve local artifact cues while learning global authenticity features, which leads to more stable forged/authentic prediction and better downstream evidence guidance.

\textbf{Effect of Difficulty Mining.}
Table~\ref{tab:abl_difficulty_mining} evaluates the effect of iterative difficulty-aware mining. Compared with the DTD baseline, the first mining round brings a clear improvement in all localization metrics, indicating that revisiting hard samples and error-centered regions effectively strengthens the localizer. The second round achieves the best performance, improving IoU by 9.53 points over the baseline, which shows that refreshing the hard sample set after model update helps the localizer focus on remaining failure cases. However, the third round leads to a performance drop, suggesting that excessive hard-sample mining may overemphasize difficult patterns and weaken distribution balance. Therefore, we use the second-round difficulty-mined localizer in the final system.

\textbf{Effect of Report-Mask Consistency Post-processing.}
As shown in Table~\ref{tab:ablation_evidence}, removing RMC post-processing noticeably lowers localization metrics but barely affects explanation scores. Since RMC only aligns report grounding boxes with the final mask and supplements missing regions without altering semantic content, this lightweight step effectively improves spatial consistency between reports and mask evidence.

\subsection{Final Challenge Submission}

We apply the complete evidence-guided system to the official hidden test set of the GenText-Forensics Challenge. Since the test annotations are withheld, we report the official leaderboard result. Our system achieved a final score of 0.638 and ranked second in the final evaluation, demonstrating the effectiveness of the proposed solution under the official challenge protocol.

\section{Conclusion}

We presented an evidence-guided detector-localizer-reasoner system for the GenText-Forensics Challenge. By combining auxiliary localization loss, difficulty-aware mining, and report-mask consistency post-processing, our system achieved 0.638 on the hidden test set and ranked second. We also summarized practical insights from building a coordinated evidence-grounded forensic pipeline.

\begin{acks}
This work was supported in part by the National Natural Science Foundation of China (NSFC) under Grant U23B2022, and in part by the Shenzhen Science and Technology Program under Grants JCYJ20250604181211016 and SYSPG20241211174032004.
\end{acks}

\balance


\bibliographystyle{ACM-Reference-Format}
\bibliography{ms}

\end{document}